\pdfoutput=1

\documentclass[11pt]{article}

\usepackage[]{EMNLP2023}
\usepackage{times}
\usepackage{latexsym}
\usepackage[T1]{fontenc}
\usepackage[utf8]{inputenc}
\usepackage{microtype}
\usepackage{inconsolata}

\usepackage[utf8]{inputenc} 
\usepackage[T1]{fontenc}    
\usepackage{hyperref}      
\usepackage{url}            
\usepackage{booktabs}       
\usepackage{amsfonts}      
\usepackage{nicefrac}     
\usepackage{microtype}     
\usepackage{xcolor}

\usepackage{inconsolata}
\usepackage{enumitem}
\usepackage{amsmath}
\usepackage{nicefrac}   
\usepackage{wrapfig}
\usepackage{setspace}
\usepackage{makecell}
\usepackage{multirow}
\usepackage{amssymb}
\usepackage{calc}
\usepackage{graphicx}
\usepackage{subfigure}
\usepackage{physics,amsmath}
\usepackage{adjustbox}

\title{Teaching Text-to-Image Models to Communicate in Dialog}

\author{Xiaowen Sun$^1$\thanks{\quad Equal contribution.} \quad Jiazhan Feng$^1$\footnotemark[1] \quad Yuxuan Wang$^2$ \quad Yuxuan Lai$^{3,4}$ \\ \quad {\bf Xingyu Shen}$^1$ \quad {\bf Dongyan Zhao}$^{1,2}$ \\
$^1$Wangxuan Institute of Computer Technology, Peking University \\
$^2$National Key Laboratory of General Artificial Intelligence, BIGAI \\ 
$^3$Engineering Research Center of Integration and Application \\ of Digital Learning Technology, Ministry of Education \\
$^4$Department of Computer Science, The Open University of China\\
\texttt{xiaowensun@stu.pku.edu.cn \quad \{fengjiazhan,shenxy,zhaody\}@pku.edu.cn} \\
\texttt{flagwyx@gmail.com \quad erutan@pku.org.cn}}

\begin{document}
\maketitle
\begin{abstract}
\textit{A picture is worth a thousand words}, thus, 
it is crucial for conversational agents to understand, perceive, and effectively respond with pictures. 
However, we find that directly employing conventional image generation techniques is inadequate for conversational agents to produce image responses effectively.
In this paper, we focus on the innovative \textbf{dialog-to-image generation} task, where the model synthesizes a high-resolution image aligned with the given dialog context as a response.
To tackle this problem, we 
design a tailored fine-tuning approach on the top of state-of-the-art text-to-image generation models to fully exploit the structural and semantic features in dialog context during image generation. Concretely, we linearize the dialog context with specific indicators to maintain the dialog structure, and employ in-domain data to alleviate the style mismatch between dialog-to-image and conventional image generation tasks. 
Empirical results on PhotoChat and MMDialog Corpus show that our approach brings consistent and remarkable improvement with 3 state-of-the-art 
pre-trained text-to-image generation backbones. 
\end{abstract}

\section{Introduction}
\begin{figure}[t]
    \centering
    \includegraphics[width=7.8cm]{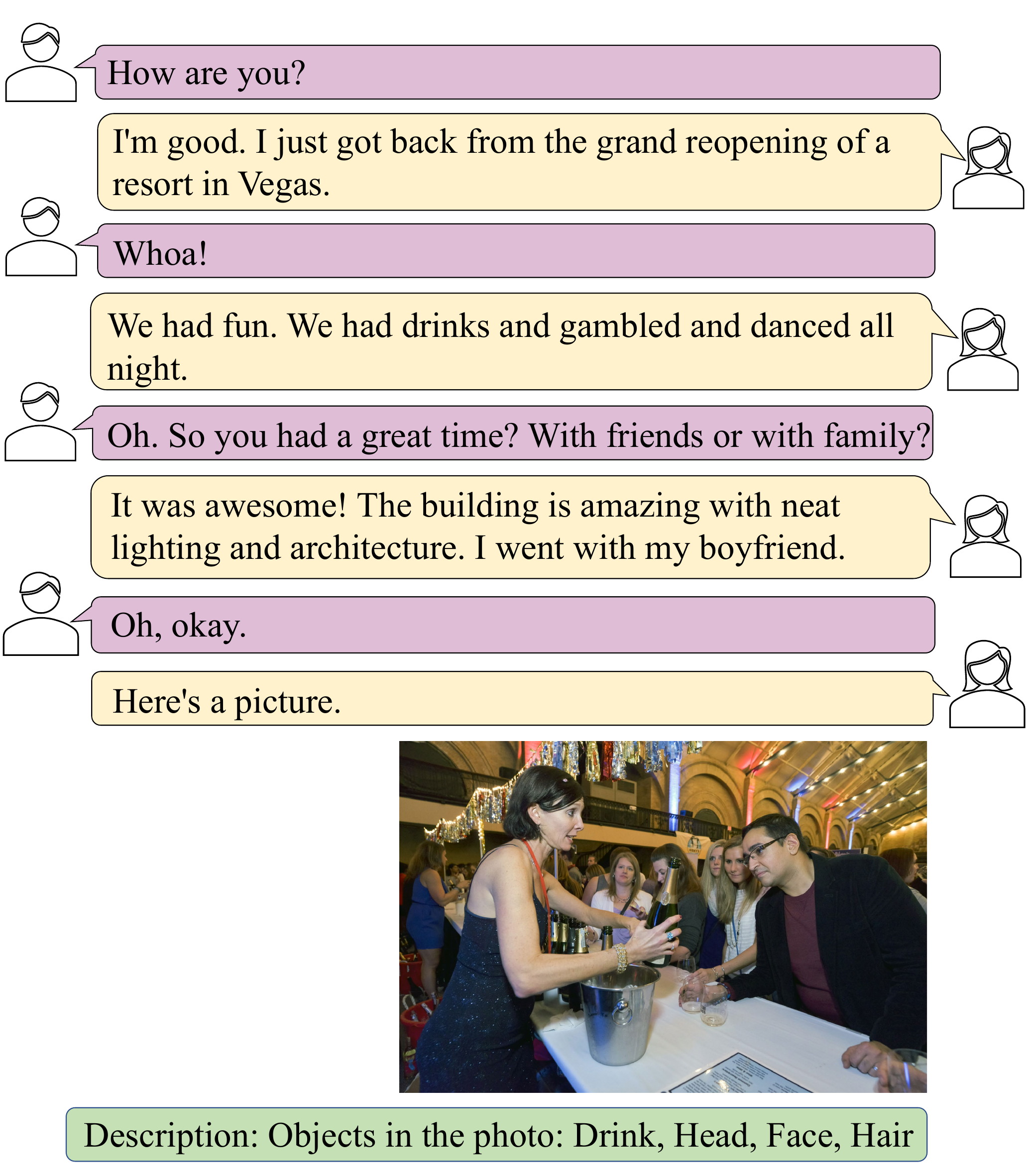}
    \caption{An example of human conversations from PhotoChat Corpus~\cite{zang-etal-2021-photochat}. 
    The speakers are talking about a grand reopening of a resort.}
    \label{fig:1}
\end{figure}

Recently, visual modalities have been playing an important role in transmitting messages. In human conversations, images can effortlessly convey a wealth of visual perception, a depth of expression that plain text often struggles to capture. Shown in Figure~\ref{fig:1}, images are necessary (i) to share more details (e.g., neat lighting and architecture) of the topic; (ii) to express the emotions (e.g., happy) about a specific event.
Therefore, it is an important capability for conversational agents to comprehend, perceive, and appropriately respond to contexts with multi-modal contents beyond mere text. 

Although some dialog models showcase remarkable capabilities of generating textual response that resembles human conversation~\cite{zhang-etal-2020-dialogpt,roller-etal-2021-recipes,ouyang2022training}, they encounter difficulties in generating images as responses.
On the other hand, in the field of image generation, various models show impressive performance on generating fascinating images according to their captions, such as DALL$\cdot$E 2~\cite{ramesh2022hierarchical}, the Latent Diffusion Model (LDM)~\cite{rombach2022high}, and UniDiffuser~\cite{bao2023one}. 

\begin{figure}[t]
    \centering
    \includegraphics[width=4.5cm]{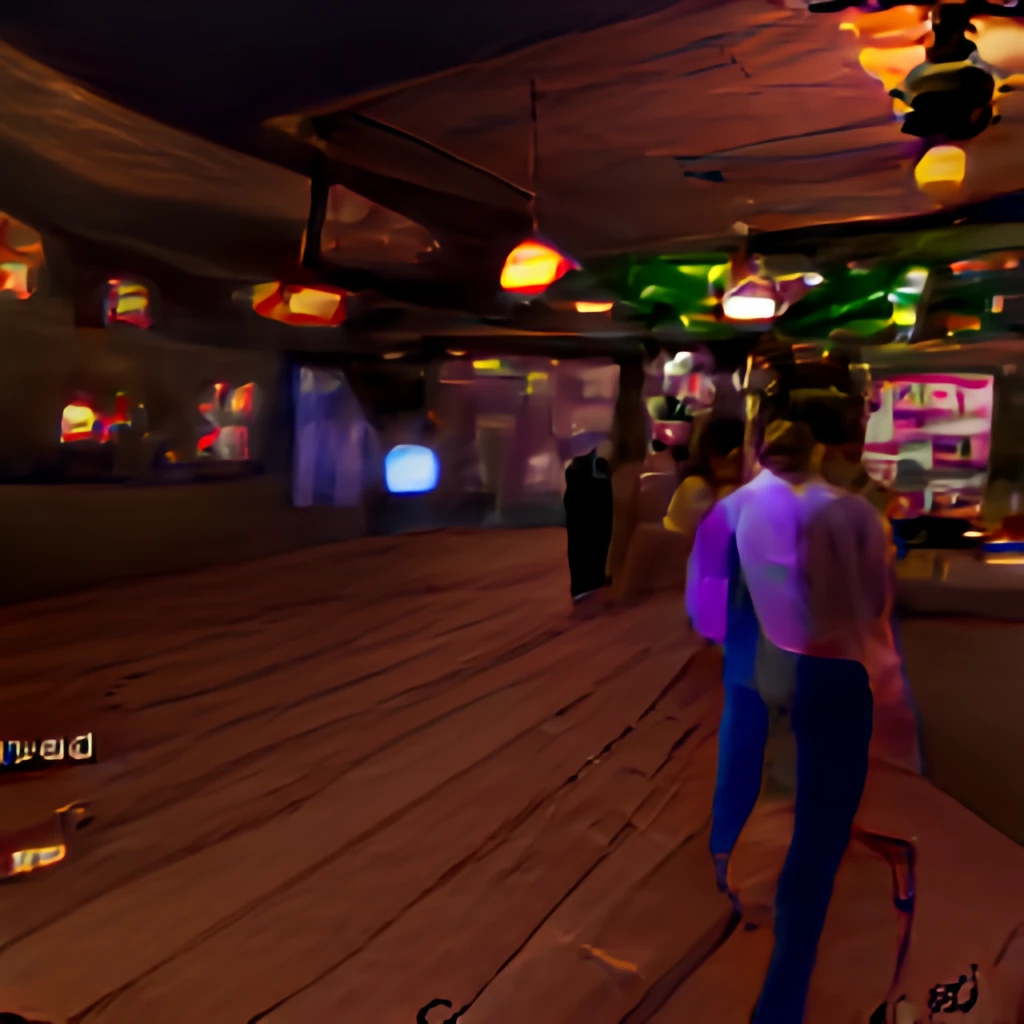}
    \caption{An image is generated by the DALL$\cdot$E with the dialog context depicted in Figure 1 serving as its input.}
    \label{fig:2}
\end{figure}

{We find it is not straightforward to exploit traditional image generation methods to equip conversational agents with the ability to generate image responses.}
{An example is Figure~\ref{fig:2}, which }
is generated by  
DALL$\cdot$E~\cite{pmlr-v139-ramesh21a}  directly conditioned on the dialog context depicted in Figure~\ref{fig:1}. 
The image does not meet the required style of real photo, and is semantically inconsistent with the dialogue context. 

To tackle this challenge, Divter~\cite{sun-etal-2022-multimodal} generates a textual image description given dialog context, and uses the generated description instead of raw context for image generation. 
However, the intermediate description is typically too brief to accurately convey the abundant image information in dialogue context.
For example, the description in Figure~\ref{fig:1} from PhotoChat conveys only 4 keywords of the object, `Drink, Head, Face, Hair', while ignoring any emotional color or details. 
In contrast, the dialog context conveys additional information including: (i) the female speaker in joyful mood through  `fun', and `danced all night'; (ii) the atmosphere of the lively party from `grand', and `with neat lighting and architecture'.  

In this paper, we focus on this novel task \textbf{dialog-to-image generation}, where given the dialog context, the model is required to generate a high-resolution 
image that is coherent with the specified conversation as a response.  
{To conquer previous problems while leveraging image generation methods, we require}
models to directly generate images from multi-modal conversational context without intermediate description translation. 
There are two main challenges:
(i) the majority of the dialogue is in real-life situations which involves numerous images of human faces, which are significantly challenging to generate; 
(ii) directly employing off-the-shelf text-to-image models for dialog-to-image generation encounters style mismatch problem.

To tackle these challenges, we improve an efficient approach for dialog-to-image generation without any intermediate translation, which maximizes the utilization 
of the semantic information in the dialog context. 
Specifically, we design a tailored fine-tuning approach on the top of state-of-the-art text-to-image generation models~\cite{bao2023all, bao2023one} to fully exploit the structural and semantic dialog context for image generation.
We linearize the dialog context with specific indicators to preserve the dialog structure during fine-tuning, and exploit in-domain data to alleviate the style mismatch challenge between dialog-to-image and conventional image generation tasks.
We argue that this carefully designed fine-tuning strategy which maximizes the utilization of knowledge acquired in pre-training could obtain extraordinary performances while keeping the computational consumptions minimal.

We conduct experiments on two datasets with three state-of-the-art image generation backends. Empirical results show that our proposed method remarkably improve the performances under all settings.
Thorough analysis demonstrates that after fine-tuning, the generated images have clearer features and richer facial expressions.
We believe this work addresses more attention to the challenging dialog-to-image generation task, and our method enlightens the usage of state-of-the-art image generation methods on specific task scenarios.

Contributions of this work are three-fold:

\begin{itemize}
    \item To the best of our knowledge, this is the first work entirely devoted to dialog-to-image generation. 
    We address the challenges of this task and argue that previous summarize-then-generation pipeline  could not make full use of the dialogue context.
    
    \item We propose a fine-tuning framework which fully utilizes state-of-the-art text-to-image models while exploring the structural and semantic information in the whole dialog context.

    \item Extensive experiments on PhotoChat and MMDialog Corpus indicate the effectiveness of our approach, where observing remarkable and consistent improvement with state-of-the-art pre-trained text-to-image generation backends.
\end{itemize}

\section{Methodology}
\textbf{Task Formulation:} In the dialogue-to-image generation task, 
given the dialog context $\mathcal{C}_{i}$, the model is required to generate a corresponding image response $\mathcal{I}_{i}$.
Different from the image captions in conventional text-to-image generation tasks, the context $\mathcal{C}_{i}=\{c_{k}\}^{K}_{k=1}$ is composed of $K$ turns of dialog, which is longer and preserving inner logical/semantic structures.
The whole dataset is constructed as dialog-image pairs: $\mathcal{D_{S}} = \{\mathcal{S}_{i}=(\mathcal{C}_{i}, \mathcal{I}_{i})\}^{n}_{i = 1}$, where $\mathcal{S}_{i}$ is an instance.

Acknowledging the commonalities between dialog-to-image generation and text-to-image generation, we have crafted a customized fine-tuning approach that takes into account the distinctive attributes of dialog text, while harnessing the capabilities of pre-existing generative models. 

\subsection{Tailored Text Concatenation Strategy}
\label{method}
Fine-tuning pre-trained text-to-image models is more effective to accomplish dialog-to-image generation building upon the model's existing ability to align text and image information. Furthermore, 
the pre-trained model has acquired a significant amount of external knowledge so that it could capture the common entities accurately, facilitating image generation. Fine-tuning instead of training from scratch also keeps computational resources and time costs minimal.

Thus we devise a customized fine-tuning approach simultaneously maintaining the unique structure of dialog text. 
Conversations, unlike image captions, typically involve multiple participants discussing varied topics. This enriches image details but demands strong text comprehension skills of the model. Inability to differentiate between speakers' information even would disrupt image generation.
Therefore, we tried various ways of connecting dialog statements (refer to \S\ref{analysis} for more details) as input and found that appending a special symbol \texttt{`\#'} before each turn of a dialog achieved the best performance among all the approaches. Concretely, given a dialog-image sample $\mathcal{S}=(\mathcal{C},\mathcal{I})$, where $\mathcal{C}$ and $\mathcal{I}$ represent the dialog context and the corresponding image, respectively. The dialog context $\mathcal{C}=\{\boldsymbol{c}_{k}\}^{K}_{k=1}$, where $\boldsymbol{c}_{k}, \forall k \in \{1,\dots,K\}$ denotes each turn before image appears.
We first add \texttt{`\#'} before $\boldsymbol{c}_{k}, \forall k \in \{1,\dots,K\}$, then concatenate all the sentences as the final text input.  

\subsection{Pre-trained Model Architecture}
Diffusion models are powerful, recently emerged deep generative models used for high-quality image generation. 
An essential element in a comprehensive generative system is a unified architecture which is capable to process various modality types as inputs. It's worth highlighting that the rise of the Transformer model~\cite{NIPS2017_3f5ee243,dosovitskiy2021an} and its utilization in generative modeling offers a promising approach to capture interactions across modalities. To ensure the quality of the generated images and facilitate the fusion of text and image information, we choose a diffusion model based on the transformer architecture as our pre-trained model. 

Followed~\citet{bao2023one}, the image encoder consists of two parts. The first part is the image autoencoder employed
in Stable Diffusion. The second part is the image CLIP~\cite{radford2021learning} (ViT-B/32). The final latent embedding for images is the concatenation of the outputs from two parts, i.e., $\boldsymbol{x}_{0}=[\boldsymbol{x}_{0}^{AE},\boldsymbol{x}_{0}^{CLIP}]$. As for the text encoder, we employ the same text CLIP as Stable Diffusion. The text CLIP outputs 77 vectors and each is 768-dimensional. We also add an extra linear layer, which reduces the dimension of each vector to 64 to obtain the final text embedding $\boldsymbol{y}_{0}$.

\begin{figure}[t]
    \centering
    \includegraphics[width=8.1cm]{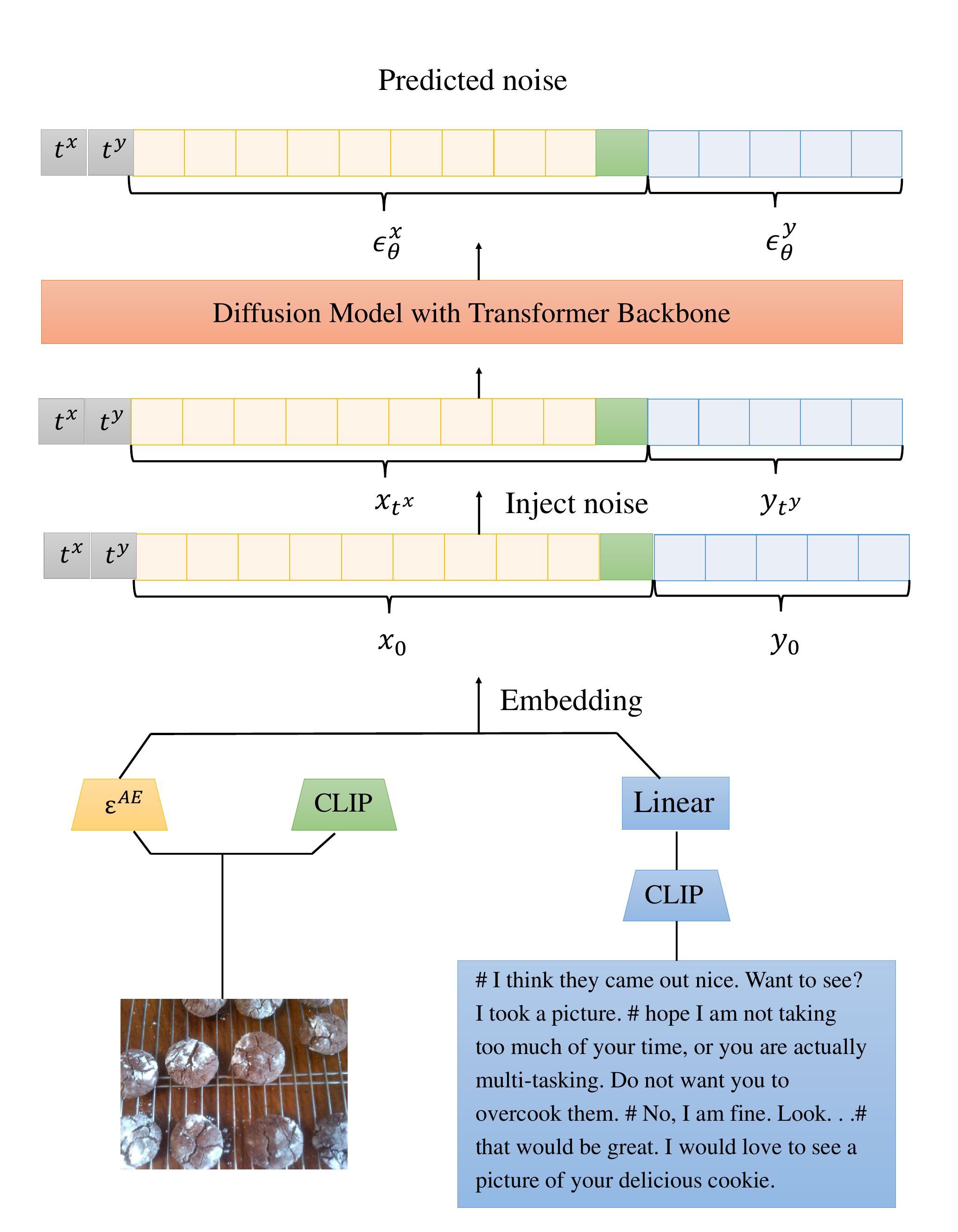}
    \caption{Implementation of the diffusion model with transformer backbone on dialog-image data.}
    \label{fig:model}
\end{figure}

We fine-tune a joint noise prediction network $\boldsymbol{\epsilon_{\theta}}(\boldsymbol{x}_{t^{x}},\boldsymbol{y}_{t^{y}},t^{x},t^{y})$ with a transformer-based backbone on the embeddings obtained above following~\citet{bao2023one}. We illustrate the model architecture in Figure~\ref{fig:model}.
We adopt the loss function mentioned in~\citet{bao2023one} which is formulated as below:
\begin{equation}
    \mathbb{E}_{\boldsymbol{x}_{0},\boldsymbol{y}_{0},\boldsymbol{\epsilon}^{x},\boldsymbol{\epsilon}^{y},t^{x},t^{y}} \|\boldsymbol{\epsilon_{\theta}}(\boldsymbol{x}_{t^{x}},\boldsymbol{y}_{t^{y}},t^{x},t^{y})-[\boldsymbol{\epsilon}^{x},\boldsymbol{\epsilon}^{y}]\|^{2}_{2},
\end{equation}
where $(\boldsymbol{x}_{0},\boldsymbol{y}_{0})$ is a random data point, [, ] denotes concatenation, $\boldsymbol{\epsilon}^{x}$ and $\boldsymbol{\epsilon}^{y}$ are sampled from standard Gaussian distributions, and $t^{x}$and $t^{y}$ are uniformly sampled from $\{1,2,\dots,T\}$ independently. During inference, we sample $\boldsymbol{x}_{0}$ conditioned on $\boldsymbol{y}_{0}$ by setting $t^{y}=0$. In Table~\ref{tab:accents}, we present the training algorithm in Algorithm 1 and the sampling procedure in Algorithm 2.
\begin{table}
\small
\centering
\begin{tabular}{l}
\toprule
\textbf{Algorithm 1} Training\\
\midrule
1: \textbf{repeat} \\
2: \quad $\boldsymbol{x}_{0},\boldsymbol{y}_{0} \sim q(\boldsymbol{x}_{0},\boldsymbol{y}_{0})$ \\
3: \quad $t^{x},t^{y} \sim$ Uniform$(\{1,2,\dots,T\})$  \\ 
4: \quad $\boldsymbol{\epsilon}^{x},\boldsymbol{\epsilon}^{y} \sim \mathcal{N}(0,\mathcal{I})$  \\ 
5: \quad Let $\boldsymbol{x}_{t^{x}}=\sqrt{\overline{\alpha}_{t^{x}}}\boldsymbol{x}_{0} + \sqrt{1 - \overline{\alpha}_{t^{x}}} \boldsymbol{\epsilon}^{x}$ \\
6: \quad Let $\boldsymbol{y}_{t^{y}}=\sqrt{\overline{\alpha}_{t^{y}}}\boldsymbol{y}_{0} + \sqrt{1 - \overline{\alpha}_{t^{y}}} \boldsymbol{\epsilon}^{y}$   \\ 
7: \quad Take gradient step on\\
\quad \quad \quad \quad \quad \quad $\nabla_{\theta}\|\boldsymbol{\epsilon_{\theta}}(\boldsymbol{x}_{t^{x}},\boldsymbol{y}_{t^{y}},t^{x},t^{y}) -[\boldsymbol{\epsilon}^{x},\boldsymbol{\epsilon}^{y}]\|^{2}_{2}$  \\
8: \textbf{until} converged \\
\bottomrule
\end{tabular}

\quad
\quad
\quad
\quad
\quad
\quad
\small
\begin{tabular}{l}
\toprule
\textbf{Algorithm 2} Sampling of $\boldsymbol{x}_{0}$ conditioned on $\boldsymbol{y}_{0}$\\
\midrule
1: $\boldsymbol{x}_{T} \sim \mathcal{N}(0,\mathcal{I})$ \\ 
2:  \textbf{for} $t=T,\dots,1$ \textbf{do} \\ 
3: \quad $\boldsymbol{z}^{x} \sim \mathcal{N}(0,\mathcal{I})$ if $t > 1$, else $\boldsymbol{z}^{x}=0$\\ 
4: \quad $\boldsymbol{x}_{t-1} = \frac{1}{\sqrt{\alpha_{t}}}\left(\boldsymbol{x}_{t} - \frac{\beta_{t}}{\sqrt{1-\overline{\alpha}_{t}}}\boldsymbol{\epsilon_{\theta}}^{x}(\boldsymbol{x}_{t},\boldsymbol{y}_{0},t,0)\right)$ \\
\quad \quad \quad \quad \quad $+\sigma_{t}\boldsymbol{z}^{x}$ \\ 
5: \textbf{end for} \\ 
6: \textbf{return} $\boldsymbol{x}_{0}$ \\ 
\bottomrule
\end{tabular}

\caption{Training and sampling algorithm.}
\label{tab:accents}
\end{table}

\section{Experiments}
\begin{table*}
\small
\centering
\setlength{\tabcolsep}{15pt}
\begin{tabular}{llll}
\toprule
\textbf{Models} & \textbf{FID $\downarrow$} &\textbf{IS $\uparrow$}  & \textbf{CLIP-I $\uparrow$} \\
\midrule
\multicolumn{4}{l}{\textbf{\textit{PhotoChat}}} \\
DALL$\cdot$E 2 & 124.10  & ~8.8 $\pm$ 0.7  & 42.18\\
Divter & \textbf{29.04} & 15.8 $\pm$ 0.6 & –  \\
\midrule
U-ViT-Small & 97.76 & 10.8 $\pm$ 0.6 & 56.61 \\
\textbf{U-ViT-Small + ours} & 86.27 (-11.49) & 12.5 $\pm$ 1.2 (+1.7) & 60.61 (+4.00) \\
\midrule
U-ViT-Small(Deep) & 96.07  & 10.9 $\pm$ 0.7  & 56.87 \\
\textbf{U-ViT-Small(Deep) + ours} & 87.59 (-8.48) & 12.4 $\pm$ 1.6 (+1.5) & 60.76 (+3.89) \\ 
\midrule
UniDiffuser-v1 &  105.42 &  11.7 $\pm$ 1.0  & 50.59  \\
\textbf{UniDiffuser-v1 + ours} & 67.93
\textbf{({-37.49})} & \textbf{17.1 $\pm$ 1.7} (\textbf{+5.4}) & \textbf{61.82} (\textbf{+10.63}) \\
\midrule
\multicolumn{4}{l}{\textbf{\textit{MMDialog}}} \\
DALL$\cdot$E 2 & - & 10.65 $\pm$ 0.28 & -\\
MiniGPT-5 & - & 19.63 & - \\
Divter & - & 20.53 $\pm$ 0.50 & - \\
\midrule
U-ViT-Small & 52.95 & 13.54 $\pm$ 0.19 & 41.79\\
\textbf{U-ViT-Small + ours} & 23.90 \textbf{(-29.05)} & 16.78 $\pm$ 0.31 (+3.24) &  43.11 (+1.32)\\
\midrule
U-ViT-Small(Deep) & 49.07 & 14.22 $\pm$ 0.43 & 43.34 \\
\textbf{U-ViT-Small(Deep) + ours} & \textbf{21.98} (-27.09) & 16.27 $\pm$ 0.41 (+2.05) & \textbf{43.68} (+0.34)\\
\midrule
UniDiffuser-v1 & 31.49 & 15.31 $\pm$ 0.29 & 33.77 \\
\textbf{UniDiffuser-v1 + ours} & 22.72 (-8.77) & \textbf{21.61 $\pm$ 0.53} (\textbf{+6.30}) & 38.56 \textbf{(+4.79)} \\
\bottomrule

\end{tabular}

\caption{Evaluation results
on the \textbf{PhotoChat} and \textbf{MMDialog} test set. Models in \textbf{bold} are equipped with our fine-tuning approach.
Numbers in (parenthesis) indicate  improvements after pre-training with our strategy.
Meanings of metrics see \S\ref{s_metrics}.
}
\label{2}
\end{table*}

\subsection{Datasets}
To evaluate the performance of our method, we conduct experiments on the PhotoChat and MMDialog dataset. 

\paragraph{PhotoChat} PhotoChat dataset~\cite{zang-etal-2021-photochat} is a multimodal conversational dataset that casts light on the photo sharing behavior in online messaging. 
Each dialog is paired with a user image that is shared during the conversation. 
There are 10,286, 1000 and 1000 instances in the training, validation and test set, respectively.  

\paragraph{MMDialog} MMDialog~\cite{feng-etal-2023-mmdialog} is the largest multi-modal conversation dataset 
containing massive topics.
We randomly select 1,000,000 instances from the three-million-scale original training set for training due to computational limits. We exploit the original validation and test set containing 23,812 and 23,772 instances, respectively.

\paragraph{Preprocessing} For both datasets, we retain the text in each turn before the image appears as the context and use the image as the target for generation.
Since we freeze the parameters in the text CLIP, which
only supports inputs with a maximum length of 77 tokens, we truncate the text from the back to the front to ensure that the length of the input text does not exceed 77 tokens.

\subsection{Baselines}
For a comprehensive evaluation of our performance in dialog-to-image generation, we conducted comparative analyses with Divter and several text-to-image generation models.

\paragraph{Divter}
~\citet{sun-etal-2022-multimodal} 
introduces a customized transformer structure for generating multimodal responses. This model demonstrates an effective capability to comprehend multi-modal dialogue contexts and produce informative textual and high-resolution image responses. 
Divter is the official baseline for MMDialog.

\paragraph{MiniGPT-5}~\citet{zheng2023minigpt}
aim to enhance the abilities of LLMs for multimodal generation through the alignment of the LLM with a pre-trained text-to-image generation model.

\paragraph{Text-to-Image Models}
We directly employ several pre-trained generative models without any additional training for dialog-to-image generation. These models all exhibit outstanding performance in text-to-image generation.

\begin{itemize}
    \item \textbf{DALL$\cdot$ E 2\footnote{\url{https://github.com/LAION-AI/dalle2-laion}}:}~\citet{ramesh2022hierarchical} propose a two-stage model, DALL$\cdot$E 2: a prior that generates a CLIP image embedding given a text caption, and a decoder that generates an image conditioned on the image embedding.
    
    \item \textbf{U-ViT\footnote{\url{https://github.com/baofff/U-ViT}}:} 
    ~\citet{bao2023all} design a ViT-based architecture (named U-ViT) for image generation with diffusion models. Followed~\citet{bao2023all}, we adopt two different configurations of U-ViT as our baselines, namely U-ViT-Small with 44M parameters and U-ViT-Small (Deep) with 58M parameters.

    \item \textbf{UniDiffuser\footnote{\url{https://github.com/thu-ml/unidiffuser}}:}~\citet{bao2023one} propose a unified framework (dubbed UniDiffuser) based on U-ViT. We utilize UniDiffuser-v1, the best version of UniDiffuser provided by~\citet{bao2023one}. UniDiffuser-v0 is trained on LAION-5B~\cite{schuhmann2022laion} at 512x512 resolution, which contains noisy webdata of text-image pairs. While 
    UniDiffuser-v1 is resumed from UniDiffuser-v0, and is further trained with a set of less noisy internal text-image pairs. 
\end{itemize}

\subsection{Implementation Details}
We initialize the  UniDiffuser model weights with pre-trained
UniDiffuser-v1\footnote{\url{https://huggingface.co/thu-ml/unidiffuser-v1}}. Similarily, we set the initial weights of the two U-ViT models using the corresponding pre-trained checkpoints.

In all fine-tuning stage, we freeze the parameters of both the image encoder and the text encoder and  only train the noise prediction network.
Among all the experiments, we use the AdamW optimizer~\cite{loshchilov2017decoupled} with a learning rate of 3e-5, a weight decay of 0.03 and running coefficients of $(\beta_{1}, \beta_{2}) = (0.9, 0.9)$. We train with mixed precision for efficiency 
and use DPM-Solver~\cite{lu2022dpm} with 50 steps in all experiments. More implementation details can be found in Appendix~\ref{sec:appendix1}.

\subsection{Evaluation Metrics}
\label{s_metrics}
We employ a diverse set of metrics to measure the fidelity and quality of generated images, including
FID~\cite{heusel2017gans}, IS~\cite{salimans2016improved}, and CLIP-I~\cite{rombach2022high}. 
FID measures the distance between generated images and real images. A smaller FID score indicates that the generated images are more closer to the ground-truth. 
IS is a measure of the clarity and diversity of generated images and larger values mean higher quality. 
The FID and IS scores are computed via \url{https://github.com/toshas/torch-fidelity}. 
CLIP-I evaluates the similarity between generated and ground-truth images and a higher metric indicates that the generated images resemble the ground-truth more closely, which is computed by
\url{https://github.com/Taited/clip-score}. 

\subsection{Experimental Results}
Table~\ref{2} 
outlines the results of our method on common text-to-image models and baselines across both PhotoChat and MMDialog test dataset, respectively. From the numbers in parenthesis in the Table~\ref{2}, it can be observed that our method has a consistent improvement among all the listed text-to-image generation models in terms of FID, IS and CLIP-I metrics on both the two datasets. 
Especially UniDiffuser-v1 proposed by~\citet{bao2023one} achieves the best performance in terms of the IS and CLIP-I score across the PhotoChat test dataset after fine-tuning. Futhermore, UniDiffuser-v1 demonstrates superior performance in IS score across the MMDialog test dataset.

We list the results of concatenating the dialog text with \texttt{`\#'} then utilizing the text-to-image generation model directly for inference. Although the models we select all perform excellently on text-to-image generation, they cannot complete dialog-to-image generation task well. This experimentally illustrates that we cannot simply consider dialog-to-image generation as text-to-image generation. Hence, the specified method for dialog-to-image generation needs to be studied separately.  

Divter~\cite{sun-etal-2022-multimodal} achieves the best FID among all the models on PhotoChat dataset. 
We think the first reason is their 12-billion-scale text-to-image translator, DALL$\cdot$E, which is 12 times larger than UniDiffuser. 
Besides, Divter incorporate 5M additional training data from ImageNet~\cite{5206848} and YFCC100M~\cite{10.1145/2812802}, while we only leverage 10,000 instances from PhotoChat training set.
Despite the significant disparity in parameter and training data scales, our methods still outperform Divter in IS and CLIP-I on PhotoChat.

Additionally,  although MiniGPT-5~\cite{zheng2023minigpt} with sophisticated structure leverages Vicuna LLM which parameter is 13 billion, our method surprisingly outperforms MiniGPT-5 on MMDialog dataset in terms of the IS score while saving computational resources and training time. This demonstrates the effectiveness of our approach in dialog-to-image generation.

Above all, the comparison results shown in Table~\ref{2}
indicate 1): our method is effective and feasible for solving dialog-to-image generation problem; 2) our method is compatible with most text-to-image generation models. With very few improvements, the initial generative model can efficiently accomplish dialog-to-image generation according to our proposed method while minimizing computational resources and time expenditure.

\section{Further Analysis}

\subsection{Impact on Concatenation Method}
\begin{table}
\small
\centering
\setlength{\tabcolsep}{3pt}
\begin{tabular}{lrc}
\toprule
\textbf{Concatenation Strategy} 
& \textbf{FID $\downarrow$} & \textbf{IS $\uparrow$}\\
\midrule
UniDiffuser-v1(` ') & 69.92 & 16.4 $\pm$ 0.9 \\
UniDiffuser-v1([PER1]\&[PER2]) & 72.93 & 14.7 $\pm$ 1.9 \\
UniDiffuser-v1(`A:'\&`B:') & 104.14 & 11.2 $\pm$ 0.8 \\
\midrule
UniDiffuser-v1(\texttt{`\#'})  & \textbf{67.93} & \textbf{17.1 $\pm$ 1.7} \\
\bottomrule
\end{tabular}
\caption{Evaluation results on PhotoChat dataset with UniDiffuser-v1 and different concatenation strategies. 
\textbf{(` ')} means that the content of each turn in dialog is separated with spaces; \textbf{([PER1]\&[PER2])} means that the speaker is explicitly indicated with special token [PER1] / [PER2] before the utterances; \textbf{(`A:'\&`B:')} means that `A:' / `B:' instead of specifical tokens are used for indicators; and \textbf{(\texttt{`\#'})} refers to our tailored text concatenation strategy (see \S\ref{method}). 
}
\label{ablation}
\end{table}

\label{analysis}
\begin{figure*}[ht!]
    \centering
    \resizebox{0.92\textwidth}{!}{
    \includegraphics[width=1\textwidth]{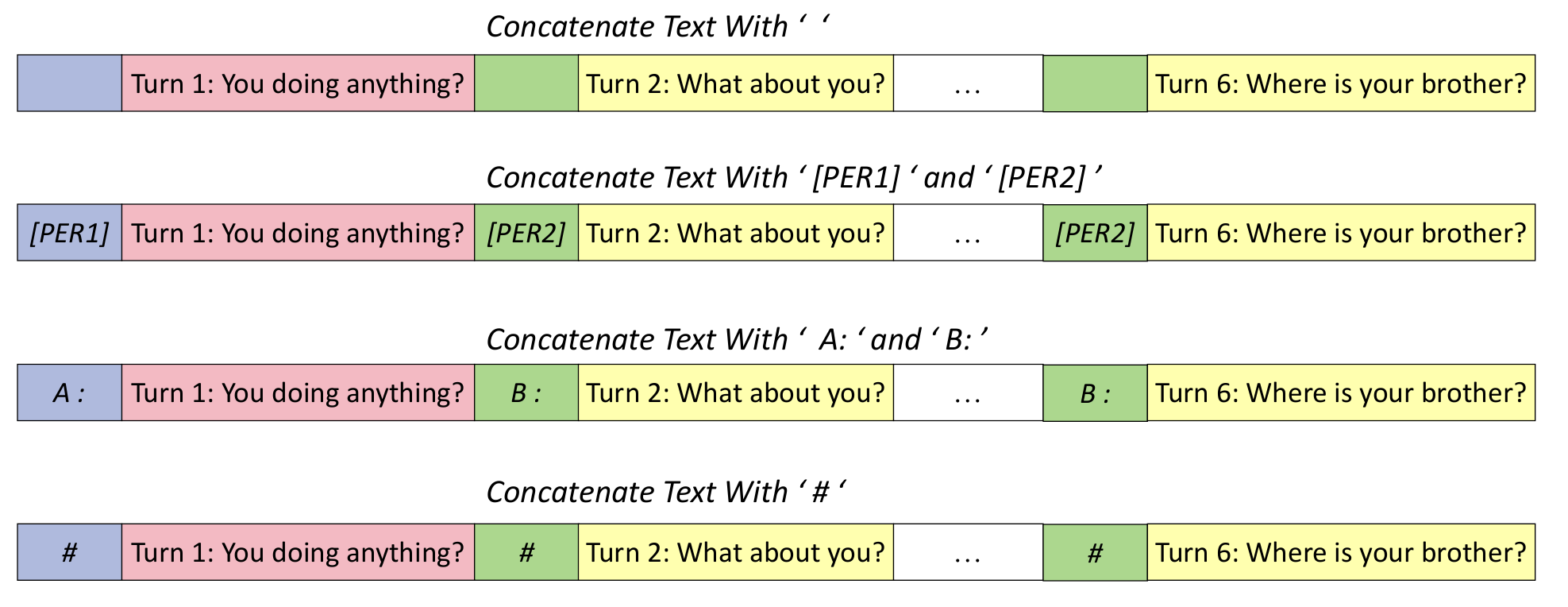}
    }
    \caption{Different ways to indicate the dialogue structures.}
    \label{concatenation}
\end{figure*}

\begin{figure}[t]
    \centering
    \includegraphics[width=7.5cm]{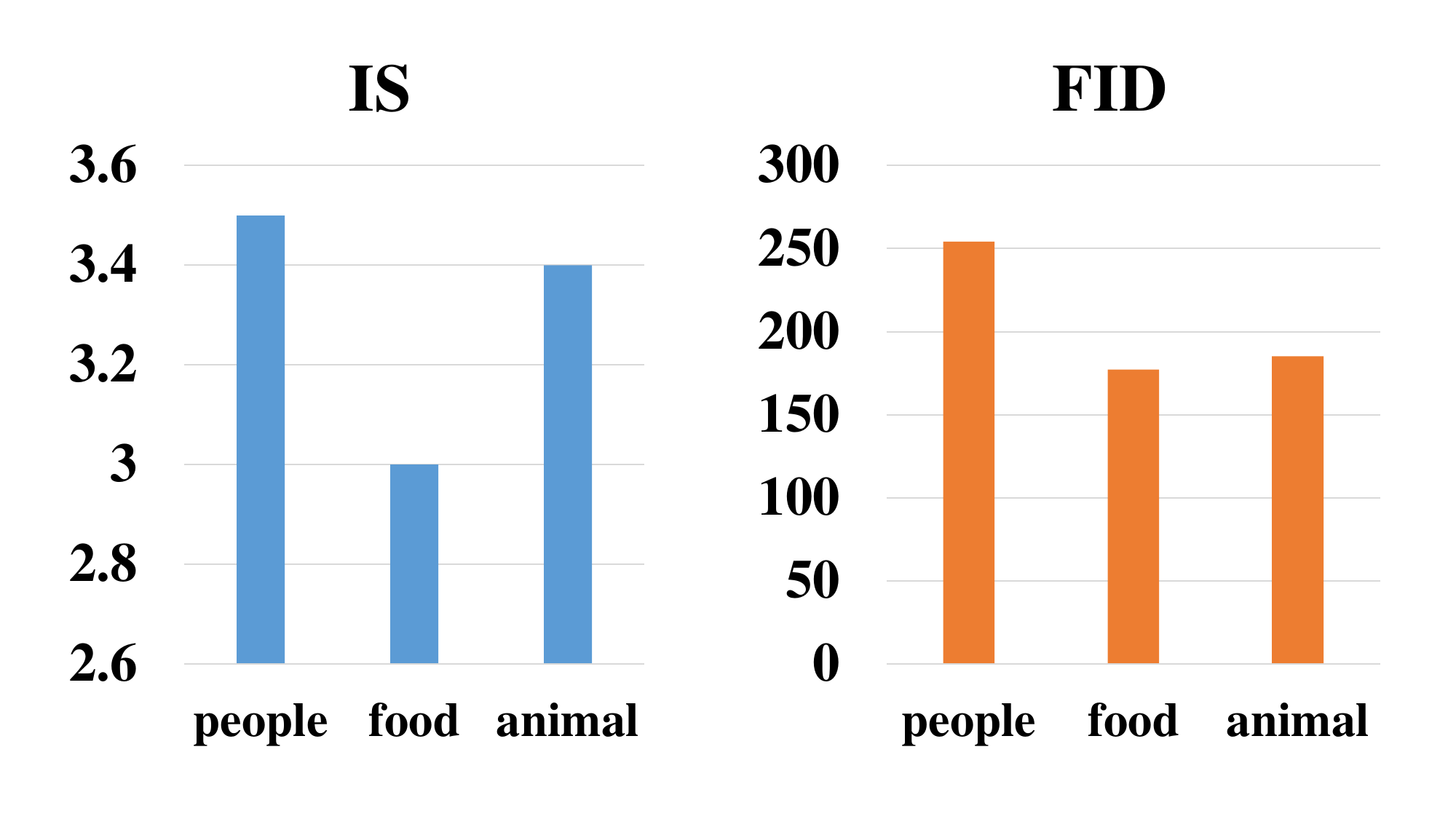}
    \caption{The IS and FID scores of 50 samples each belonging to the categories of people, food, and animals from PhotoChat test set.}
    \label{fig:cate}
\end{figure}
We tried several different fine-tuning approaches according to the distinctive attributes of dialog text in Table~\ref{ablation}. Figure~\ref{concatenation} clearly demonstrates these four methods.

As depicted in Table~\ref{ablation}, concatenating the sentence with \texttt{`\#'} achieves the best performance. We believe there are several reasons as follows: (i) given that the clip tokenizer's vocabulary lacks the two special token [PER1] or [PER2] and we freeze the parameters of the pre-trained text clip, the model is unable to distinguish between different interlocutors as our expectation; (ii) we speculate that the reason for the poor performance of (`A:'\&`B:') is that the model probably confuses the special token and the article `a'.

\subsection{Case Study}
\begin{table*}
\small
\centering
\begin{tabular}{l}
\toprule
\textbf{Case Study} \\
\midrule
\textbf{\textit{dialog context}}\\
\textbf{A:} Hello! How have you been?  \\
\textbf{B:} Hey! I'm doing okay. We missed you at dinner last night.\\
\textbf{A:} Yeah, so sorry. I couldn't get out of work.  Maybe we can make plans for next week? \\
\textbf{B:} I understand that. It was a nice excuse to get dressed up after so many weeks locked inside.  \\
\textbf{A:} Wish I could have been there! I need to get out for sure.\\
\textbf{B:} Uncle Alessandro even came out! I hadn't seen him in months. \\
\textbf{A:} Aw, how is he doing? \\
\textbf{B:} He's good. Still as well dressed and nice as ever. He's only in town for a few weeks.  \\
\textbf{A:} Haha, he always had good style. \\
\textbf{B:} \\
\midrule
\textbf{\textit{image responses}}\\
\quad \begin{minipage}[b]{0.4\columnwidth}
    \centering
    {\includegraphics[width=\textwidth]{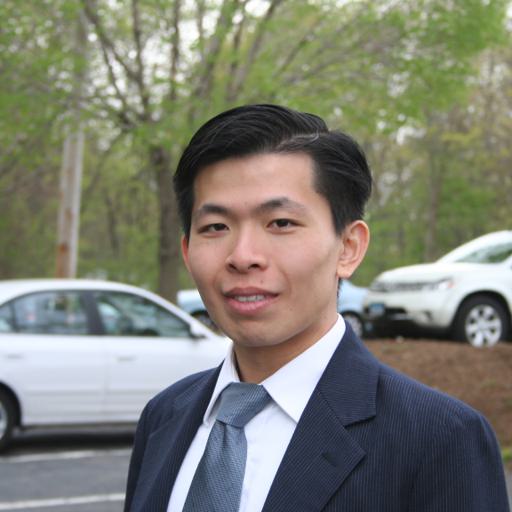} \captionsetup{font=small} \caption*{the ground truth}}
\end{minipage}
\quad \quad \begin{minipage}[b]{0.4\columnwidth}
    \centering
    {\includegraphics[width=\textwidth]{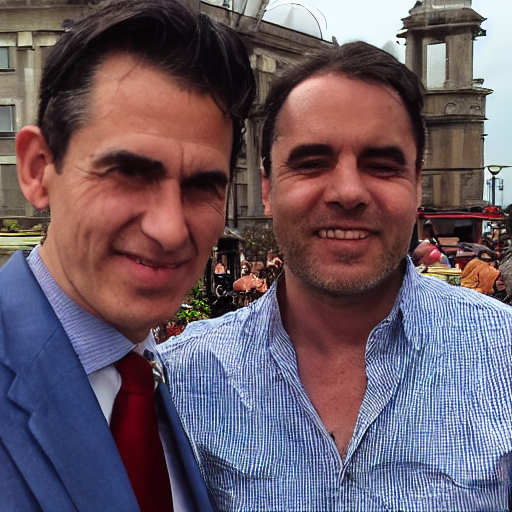} \captionsetup{font=small} \caption*{ \textbf{UniDiffuser-v1+ours}}}
\end{minipage} 

\quad \quad \begin{minipage}[b]{0.4\columnwidth}
    \centering
    {\includegraphics[width=\textwidth]{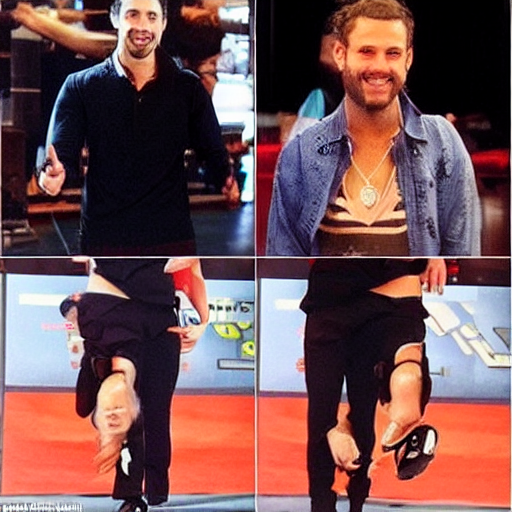} \captionsetup{font=small} \caption*{original UniDiffuser-v1}}
\end{minipage} \\

\bottomrule
\end{tabular}
\caption{A case study in PhotoChat test set. 
}
\label{case}
\end{table*}

To further investigate the quality of images generated by models after fine-tuning according to our proposed method, we show an example on the PhotoChat test dataset in Table~\ref{case}. The given context is about `Uncle Alessandro'. As we can see, UniDiffuser-v1 after fine-tuning can generate a realistic high-resolution image containing clear human face which is coherent to the background, forming a striking contrast with the strange image generated by it without any training. The high-quality generated image is comparable to that real-world ground truth, which demonstrates the practicability of our method. More qualitative examples can be found in Appendix~\ref{appendix2}.

\subsection{Performance on Dialog Categories}

We randomly select 50 samples each belonging to the categories of people, food, and animals from PhotoChat test set and calculate the FID and IS for the images in these three categories generated by UniDiffuser after fine-tuning. As illustrated in Figure~\ref{fig:cate}, the images of people have the highest FID while the images of food have the lowest both FID. This is within our expectation since people exhibit a wide range of body language and facial expressions, which is difficult for models to accurately generate consistent images. The model that has been trained on the text-to-image dataset can generate relatively simpler food and animal more accurately since it has sufficient external knowledge. We also find that the images of people have the highest IS while the images of food have the lowest IS.
\section{Related Works}

\paragraph{Multi-Modal Dialog Models: }
Numerous advanced contributions have emerged in parallel with the evolution of multi-modal dialogue datasets~\cite{das2017visual,mostafazadeh2017image,shuster2018image,zang-etal-2021-photochat,liao2021mmconv,feng-etal-2023-mmdialog}. Several efforts have been undertaken to enhance the performance of conversational agents in image-grounded dialogues through various dialogue modeling approaches~\cite{qi2020two,lee2021constructing}. Researchers~\cite{yang2021open,liang2021maria} investigate enhancing the textual representations of generated dialogue responses using associative visual scenes.~\citet{zang-etal-2021-photochat} suggest two objectives: one involves predicting the intention to share a photo in the next dialogue turn, and the other is a dialogue-based image retrieval task for finding the most suitable photo based on the conversation context. Additionally, they introduce a dual-encoder model that leverages object labels to encode image characteristics. Nevertheless, the effectiveness of the retrieval-based approach is constrained in particular domains due to the constraints posed by the size of the pre-established conversational history database. This is particularly notable for less common or specialized contexts not accounted for in the history, where the array of image responses in a retrieval system remains constant. In a recent development,~\citet{sun-etal-2022-multimodal} have pioneered the creation of a multi-modal dialogue response generation model called Divter.  However, it has not yet moved beyond the conventional approach of traditional text-to-image generation, which commonly employs brief captions to generate images.~\citet{zheng2023minigpt} present an innovative interleaved vision-and-language generation approach that incorporates the concept of "generative vokens" to serve as the bridge for harmonized image-text outputs. Nevertheless MiniGPT-5 still utilize the caption feature from the text encoder of Stable Diffusion 2.1 in the first stage training. Therefore, we explore a tailored method to directly generate images from conversational information without any intermediate translation.

\paragraph{Text-to-image Generation:}
In the research of text-to-image generation, various studies have been extensively
explored.~\citet{mansimov2015generating} show the Draw generative model~\cite{pmlr-v37-gregor15} is capable to generate images from natural language descriptions.~\citet{pmlr-v48-reed16}  introduce a generative adversarial network to enhance the image's fidelity. Subsequently, several enhancement techniques persist in fine-tuning the generation architecture, such as stacked generators~\cite{zhang2017stackgan}, attentional network~\cite{xu2018attngan}, and extra knowledge~\cite{li2019object}.~\citet{nguyen2017plug} propose a unified probabilistic interpretation of related activation maximization methods
to produce high-quality images.~\citet{cho-etal-2020-x} apply consistent masking using a wide range of masking ratios and matched the appropriate pre-training datasets with the relevant objectives.~\citet{pmlr-v139-ramesh21a} and~\citet{ding2021cogview} employ transformer-based techniques that autoregressively model the text and image tokens as a single stream of data. Recently, diffusion models have been employed to address text-to-image generation tasks due to their flexibility and strength. The Latent Diffusion Model (LDM)~\cite{rombach2022high} enables conditional image generation while streamlining the training and sampling processes for denoising diffusion models without sacrificing quality.~\citet{saharia2022photorealistic} present Imagen, a text-to-image diffusion model that achieves an unparalleled level of photorealism and a profound understanding of language. Although these various models show impressive performance on text-to-image generation, they fail to complete dialog-to-image-generation task. Therefore, we design a tailored fine-tune approach in consideration of the characteristics of dialog text to enable models to generate images based on dialog.  

\section{Conclusions}
In this paper, we highlight a new problem: dialog-to-image generation. 
We firstly address the challenges of this task by showing that directly use text-to-image models on the top of dialogue context or generated image descriptions could not fully exploit the semantic and structural information in the dialog context. 
To tackle this problem, we present an effective fine-tuning based strategy with state-of-the-art text-to-image models as backbones.
Extensive experiments across PhotoChat and MMDialog datasets on three backbone model demonstrate  
remarkable and consistent improvement brought by our approach.
Thorough analysis shows that our methods help the model in capturing the semantic and structural features from the dialog context and generating the images in expected style with richer facial expressions. 

\section*{Limitations}
Besides its merits, this work still has limitations that could be further explored.  
On the one hand, due to the lack of reliable and secure open-source code for some text-to-image generation models, we are unable to verify whether our method is effective for them. If these models become open-source in the future, we will further validate whether our approach is efficient. On the other hand, we only explore to utilize pure text contexts to generate images. In future work, we aim to further investigate to generate images based on various modalities, such as images and videos.

\section*{Ethics Statement}
This paper proposes an effective method to tackle dialog-to-image generation problem. We acquire all datasets and codes for models though academic access. 
Thus, there will not be any ethical problems or negative social consequences from the research. The proposed method does not introduce ethical/social bias in the data.

\bibliography{custom}
\bibliographystyle{acl_natbib}

\newpage~\newpage
\appendix
\section{Implementation Details}
\label{sec:appendix1}
We list more training details such as image resolution, fine-tuning batch size, total fine-tuning steps and training devices in different experiments in Table~\ref{training}.

\begin{table*}
\small
\centering
\begin{tabular}{l|ccc}
\toprule
\textbf{Models} & \textbf{Image Resolution} & \textbf{Batch Size}    &\textbf{Training Devices} \\
\midrule
\multicolumn{4}{l}{\textit{\textbf{PhotoChat}}} \\
U-ViT-Small &  256 x 256 & 740  & 1x A800-80GB GPUs \\
U-ViT-Small(Deep) & 256 x 256 & 512  & 1x A800-80GB GPUs \\
UniDiffuser-v1 & 512 x 512 & 260 &  2x A800-80GB GPUs \\
\midrule
\multicolumn{4}{l}{\textit{\textbf{MMDialog}}} \\
U-ViT-Small &  256 x 256 & 4096  & 8x A100-80GB GPUs \\
U-ViT-Small(Deep) & 256 x 256 & 4096  & 8x A100-80GB GPUs \\
UniDiffuser-v1 & 512 x 512 & 800  &  2x A100-80GB GPUs \\
\bottomrule
\end{tabular}
\caption{More training details in different experiments. 
}
\label{training}
\end{table*}

\section{More Qualitative Examples}
\label{appendix2}
\begin{table*}
\small
\centering
\begin{tabular}{l|l}
\toprule
\textbf{Example 1} & \textbf{Example 2} \\
\midrule
\multicolumn{2}{l}{\textbf{\textit{dialog context}}}\\
\textbf{A:} How are you doing? & \textbf{A:} Yesterday I went to a zoo. I saw some amazing birds \\
\textbf{B:} I'm doing good! How about you? &  \quad  and animals there. Friend..Are you there? \\
\textbf{A:} Great thanks! What are you up to tonight? & \quad There was a bird calling Magpie. It is really very pretty.  \\
\textbf{B:} Oh I just finished dinner at a restaurant with my friends.&  \textbf{B:} Yeah I am here. \\
\quad  It's down the street from my place. & \textbf{A:} Do you want to see that picture? I will send you. \\
\textbf{B:} Cool. What kind of food? & \quad  I clicked this photo. Actually its not opening.   \\ 
\textbf{A:} It's kind of a fast food with baked goods restaurant. & \quad I can not send it right now.  \\
\textbf{B:} I see. That sounds delicious. You have a photo of your  & \quad Marabou stork is the name of that bird.   \\
\quad dinner? &  \\
\textbf{A:} I got a submarine sandwich. I have picture.  &   \\
\midrule
\multicolumn{2}{l}{\textbf{\textit{image responses}}}\\
\quad \begin{minipage}[b]{0.26\columnwidth}
    \centering
    {\includegraphics[width=\textwidth]{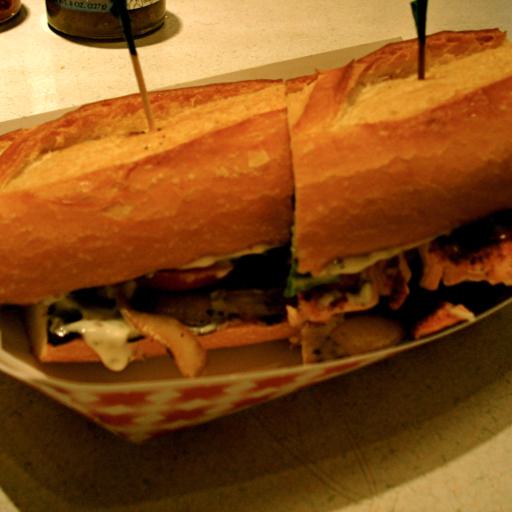}\captionsetup{font=small} \caption*{the ground truth}}
\end{minipage}
&
\quad \begin{minipage}[b]{0.26\columnwidth}
    \centering
    {\includegraphics[width=\textwidth]{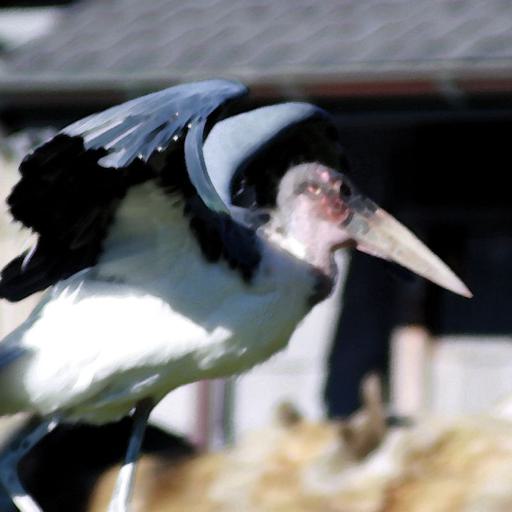} \captionsetup{font=small} \caption*{the ground truth}}
\end{minipage} \\
\midrule
\quad \begin{minipage}[b]{0.26\columnwidth}
    \centering
    {\includegraphics[width=\textwidth]{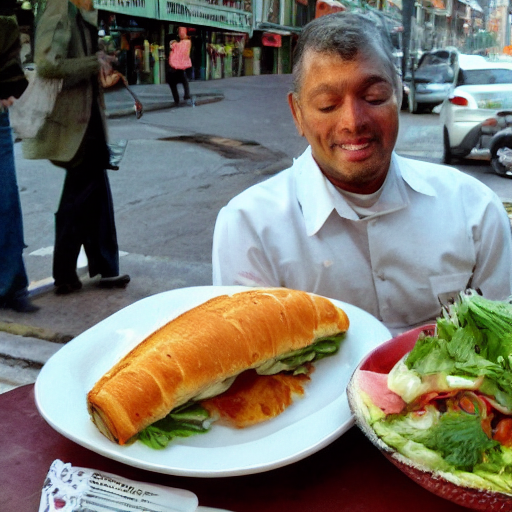} \captionsetup{font=small} \caption*{UniDiffuser-v1+ours}}
\end{minipage} 
\quad
\begin{minipage}[b]{0.26\columnwidth}
    \centering
    {\includegraphics[width=\textwidth]{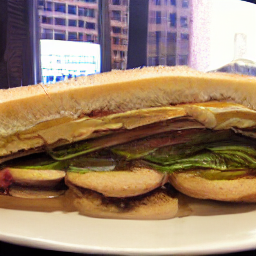}\captionsetup{font=small} \caption*{U-ViT-Small(Deep)+ours}}
\end{minipage}
\quad
\begin{minipage}[b]{0.26\columnwidth}
    \centering
    {\includegraphics[width=\textwidth]{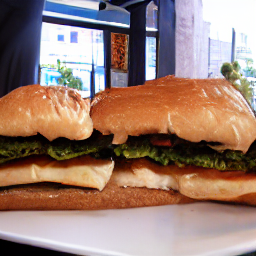}\captionsetup{font=small} \caption*{U-ViT-Small+ours}}
\end{minipage}
 & 

\quad \begin{minipage}[b]{0.26\columnwidth}
    \centering
    {\includegraphics[width=\textwidth]{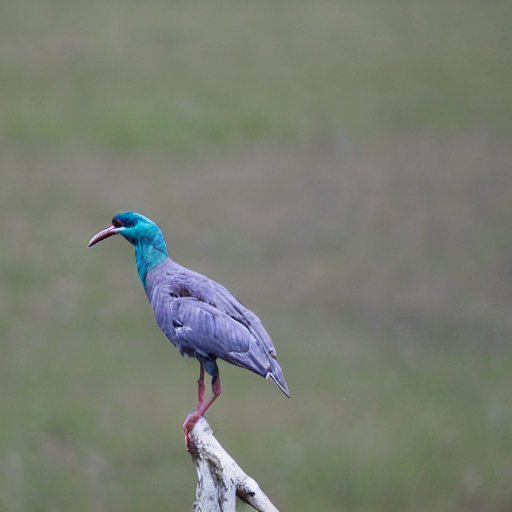} \captionsetup{font=small} \caption*{UniDiffuser-v1+ours}}
\end{minipage}
\quad
\begin{minipage}[b]{0.26\columnwidth}
    \centering
    {\includegraphics[width=\textwidth]{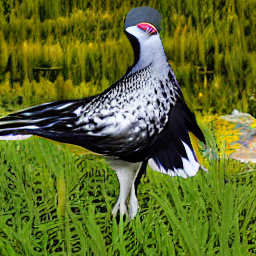} \captionsetup{font=small} \caption*{U-ViT-Small(Deep)+ours}}
\end{minipage}
\quad
\begin{minipage}[b]{0.26\columnwidth}
    \centering
    {\includegraphics[width=\textwidth]{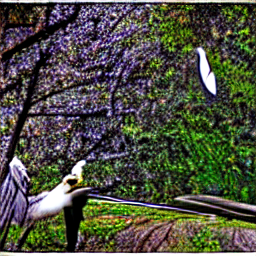} \captionsetup{font=small} \caption*{U-ViT-Small+ours}}
\end{minipage}
\\
\midrule
\quad \begin{minipage}[b]{0.26\columnwidth}
    \centering
    {\includegraphics[width=\textwidth]{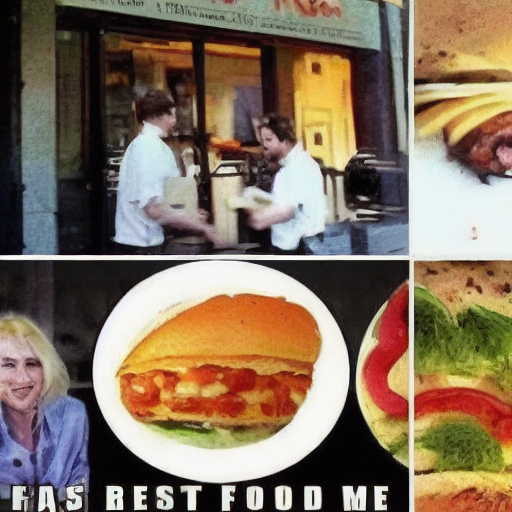} \captionsetup{font=small} \caption*{original UniDiffuser-v1}}
\end{minipage} 
\quad
\begin{minipage}[b]{0.26\columnwidth}
    \centering
    {\includegraphics[width=\textwidth]{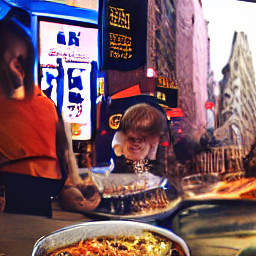} \captionsetup{font=small} \caption*{original U-ViT-Small(Deep)}}
\end{minipage}
\quad
\begin{minipage}[b]{0.26\columnwidth}
    \centering
    {\includegraphics[width=\textwidth]{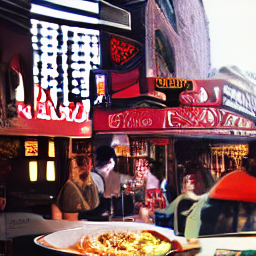} \captionsetup{font=small} \caption*{original U-ViT-Small}}
\end{minipage}
 & 

\quad \begin{minipage}[b]{0.26\columnwidth}
    \centering
    {\includegraphics[width=\textwidth]{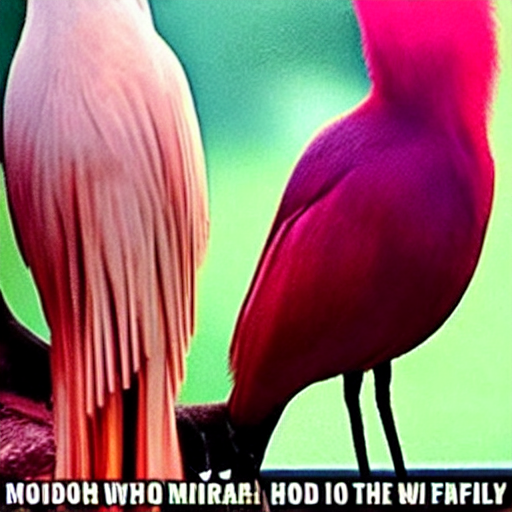} \captionsetup{font=small} \caption*{original UniDiffuser-v1}}
\end{minipage}
\quad
\begin{minipage}[b]{0.26\columnwidth}
    \centering
    {\includegraphics[width=\textwidth]{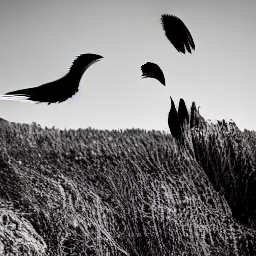} \captionsetup{font=small} \caption*{original U-ViT-Small(Deep)}}
\end{minipage}
\quad
\begin{minipage}[b]{0.26\columnwidth}
    \centering
    {\includegraphics[width=\textwidth]{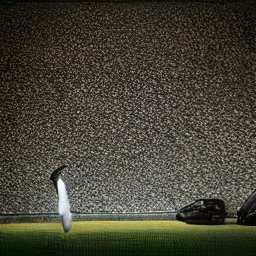} \captionsetup{font=small} \caption*{original U-ViT-Small}}
\end{minipage}
\\
\bottomrule
\end{tabular}
\caption{Examples of PhotoChat test set. 
}
\label{case study}
\end{table*}

\begin{table*}
\small
\centering
\begin{tabular}{l|l}
\toprule
\textbf{Example 1} & \textbf{Example 2} \\
\midrule
\multicolumn{2}{l}{\textbf{\textit{dialog context}}}\\
\textbf{A:} Grace is finding a waterfall when you are only looking for  & \textbf{A:} Anyone want to have a sweepstake on my US tour\\
\quad a stream. By sitting quietly in front of a waterfall, we will   &  \quad   weight gain? I'm there for 5 weeks. Winner gets one of  \\
\quad feel enriched and enlightened. The sight, sound, and power   &  \quad  my backup string winders.\\
\quad of falling water will give us a hidden message. & \textbf{B:} If you visit Chicago, I suggest Italian beef sandwiches;  \\
 & \quad deep-dish pizza, exclusive to my fair city. \\
\midrule
\multicolumn{2}{l}{\textbf{\textit{image responses}}} \\
\quad \begin{minipage}[b]{0.26\columnwidth}
    \centering
    {\includegraphics[width=\textwidth]{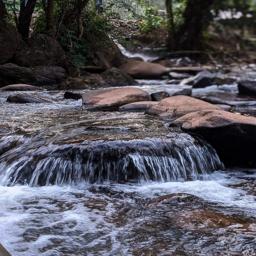}\captionsetup{font=small} \caption*{the ground truth}}
\end{minipage}
&
\quad \begin{minipage}[b]{0.26\columnwidth}
    \centering
    {\includegraphics[width=\textwidth]{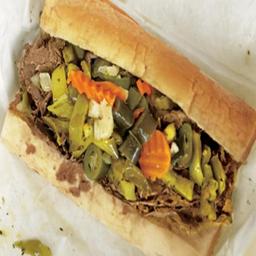}\captionsetup{font=small} \caption*{the ground truth}}
\end{minipage} \\
\midrule
\quad \begin{minipage}[b]{0.26\columnwidth}
    \centering
    {\includegraphics[width=\textwidth]{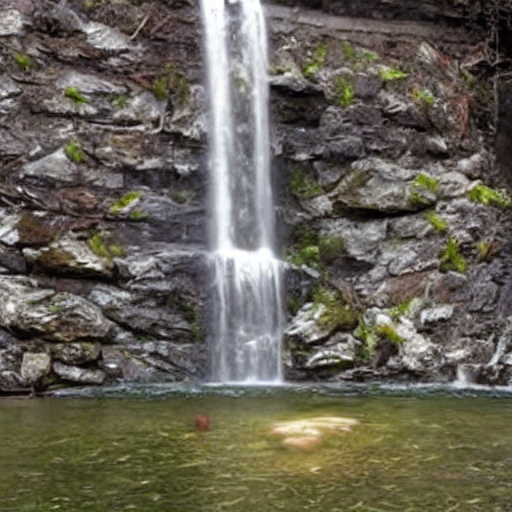}
    \captionsetup{font=small} \caption*{UniDiffuser-v1+ours}}
\end{minipage} 
\quad
\begin{minipage}[b]{0.26\columnwidth}
    \centering
    {\includegraphics[width=\textwidth]{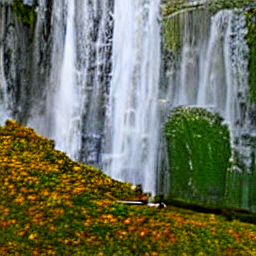}
    \captionsetup{font=small} \caption*{U-ViT-Small(Deep)+ours}}
\end{minipage}
\quad
\begin{minipage}[b]{0.26\columnwidth}
    \centering
    {\includegraphics[width=\textwidth]{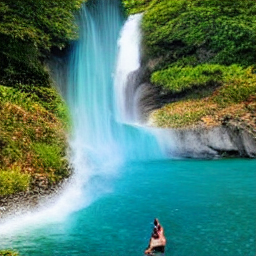}
    \captionsetup{font=small} \caption*{U-ViT-Small+ours}}
\end{minipage}
 & 

\quad \begin{minipage}[b]{0.26\columnwidth}
    \centering
    {\includegraphics[width=\textwidth]{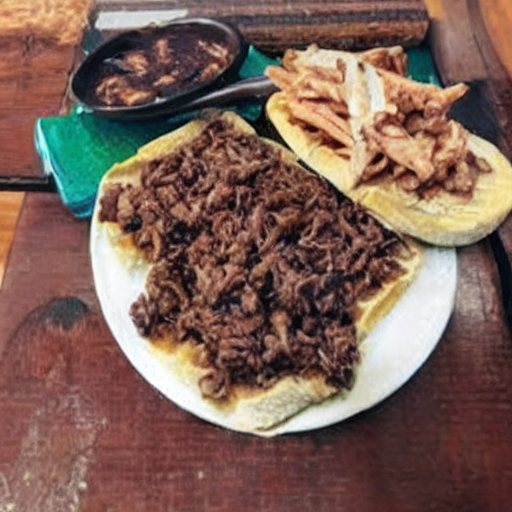}
    \captionsetup{font=small} \caption*{UniDiffuser-v1+ours}}
\end{minipage}
\quad
\begin{minipage}[b]{0.26\columnwidth}
    \centering
    {\includegraphics[width=\textwidth]{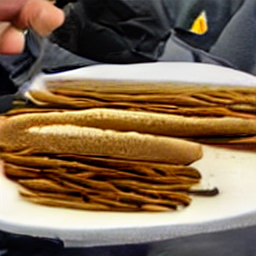}
    \captionsetup{font=small} \caption*{U-ViT-Small(Deep)+ours}}
\end{minipage}
\quad
\begin{minipage}[b]{0.26\columnwidth}
    \centering
    {\includegraphics[width=\textwidth]{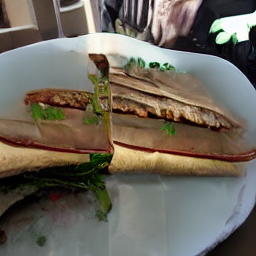}
    \captionsetup{font=small} \caption*{U-ViT-Small+ours}}
\end{minipage}
\\
\midrule
\quad \begin{minipage}[b]{0.26\columnwidth}
    \centering
    {\includegraphics[width=\textwidth]{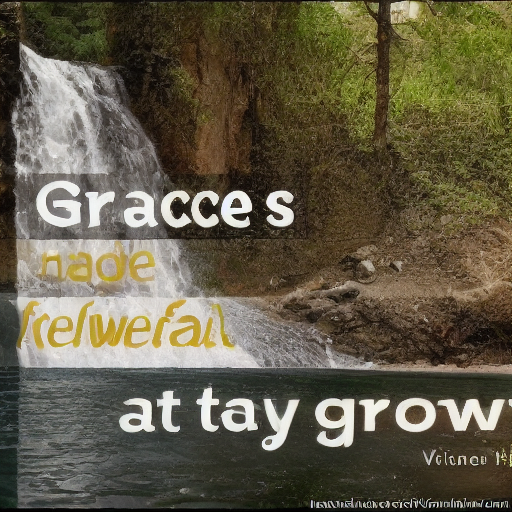}
    \captionsetup{font=small} \caption*{original UniDiffuser-v1}}
\end{minipage} 
\quad
\begin{minipage}[b]{0.26\columnwidth}
    \centering
    {\includegraphics[width=\textwidth]{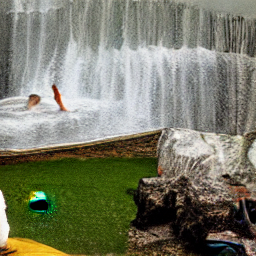}
    \captionsetup{font=small} \caption*{original U-ViT-Small(Deep)}}
\end{minipage}
\quad
\begin{minipage}[b]{0.26\columnwidth}
    \centering
    {\includegraphics[width=\textwidth]{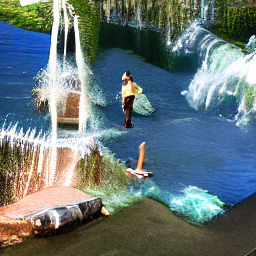}
    \captionsetup{font=small} \caption*{original U-ViT-Small}}
\end{minipage}
 & 

\quad \begin{minipage}[b]{0.26\columnwidth}
    \centering
    {\includegraphics[width=\textwidth]{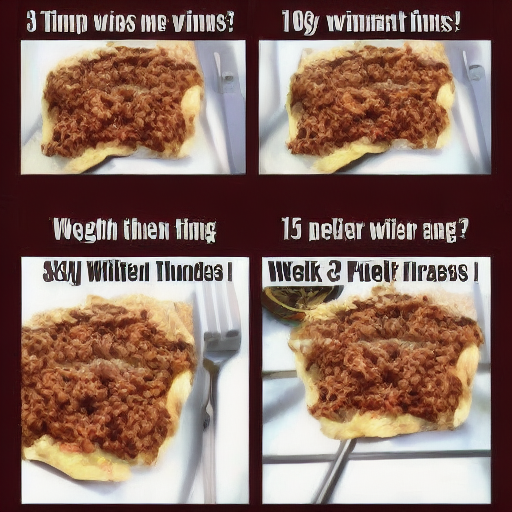}
    \captionsetup{font=small} \caption*{original UniDiffuser-v1}}
\end{minipage}
\quad
\begin{minipage}[b]{0.26\columnwidth}
    \centering
    {\includegraphics[width=\textwidth]{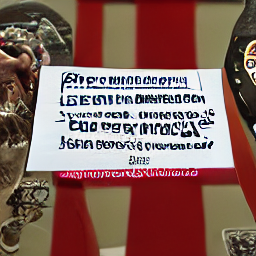}
    \captionsetup{font=small} \caption*{original U-ViT-Small(Deep)}}
\end{minipage}
\quad
\begin{minipage}[b]{0.26\columnwidth}
    \centering
    {\includegraphics[width=\textwidth]{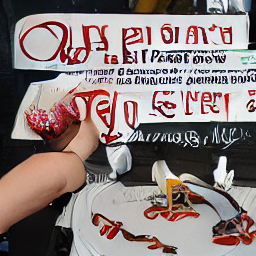}
    \captionsetup{font=small} \caption*{original U-ViT-Small}}
\end{minipage}
\\
\bottomrule
\end{tabular}
\caption{Examples of MMDialog test set. 
}
\label{case study2}
\end{table*}

Below we present more samples generated by the listed text-to-image generation models before and after fine-tuning in Tables~\ref{case study} and~\ref{case study2}.

\end{document}